**Title:**

Estimation of high frequency nutrient concentrations from water quality surrogates using machine learning methods.

**Author names and affiliations:**

María Castrillo* and Álvaro López García

Instituto de Física de Cantabria (CSIC-UC) Santander. Spain.

castrillo@ifca.unican.es, aloga@ifca.unican.es

*Corresponding author

**Abstract**

Continuous high frequency water quality monitoring is becoming a critical task to support water management. Despite the advancements in sensor technologies, certain variables cannot be easily and/or economically monitored in-situ and in real time. In these cases, surrogate measures can be used to make estimations by means of data-driven models. In this work, variables that are commonly measured in-situ are used as surrogates to estimate the concentrations of nutrients in a rural catchment and in an urban one, making use of machine learning models, specifically Random Forests. The results are compared with those of linear modelling using the same number of surrogates, obtaining a reduction in the Root Mean Squared Error (RMSE) of up to 60.1%. The profit from including up to seven surrogate sensors was computed, concluding that adding more than 4 and 5 sensors in each of the catchments respectively was not worthy in terms of error improvement.

**Keywords**

# 1. Introduction

Waterbodies must maintain a good ecological and chemical status in order to protect human health, preserve water supply and safeguard natural ecosystems and biodiversity. The assessment of the ecological status of these waterbodies in a coherent and comprehensive way would benefit from improving water quality monitoring programmes (Voulvoulis et al., 2017). To date, many substances like major nutrients (nitrogen (N) and phosphorus (P)) are mostly monitored by means of analytical discrete campaigns with low sampling frequency. Nutrient monitoring is of great importance to reduce the risk of eutrophication, a water quality problem that leads to numerous negative impacts like public health issues, fish mortality and unhealthy ecosystems, among others. Eutrophication modelling and forecasting would be improved with the collection of in-situ data on a real-time basis that could be assimilated into dynamic models.

Grab sampling has numerous drawbacks like delays in data acquisition, high costs and difficulty to detect events that take place during the time interval between two samplings. The convenience of using on-line, reliable and cost-effective sensors to monitor water quality has been widely recognized (Wade et al., 2012; Viviano et al., 2014; Valkama and Ruth, 2017), while low frequency monitoring can lead to the underestimation of risks (Brack et al., 2017).

Current commercially available sensor technologies for nutrient measurement can be classified in ion-selective electrodes (ISE), wet-chemical analysers and optical sensors. Whereas ISE are inexpensive and easy-to-use, they are criticised as inaccurate and prone to interferences and drifting. On the contrary, wet chemical analysers and optical sensors show higher resolution and accuracy, but are expensive and have high maintenance requirements and costs (Pellerin et al., 2016). These drawbacks may restrict their use for long-term in-situ and on-line outdoor monitoring. In the absence of monitoring data, modelled data can be used to obtain a more comprehensive overview of the spatial and temporal variability of the stressors and impacts on the ecosystems (Brack et al., 2017).

Traditional models used to compute these data, like watershed models or effluent discharge models, depend on complex relationships and usually require numerous input parameters and they do not usually allow obtaining high frequency data series because of the difficulty to reproduce acute events. An alternative to these models are data mining methods and machine learning. They allow exploiting the relationship between a target variable and other parameters or surrogates that can be easily and continuously measured with a high time resolution. The software that processes the available surrogate data by means of models and allows inferring the target variable(s) is called a *soft-sensor*. The implementation of soft-sensors based on surrogates allows using on-line sensors, whose convenience in comparison with time-consuming and expensive off-line chemical analysis in the laboratory has been recently recognized (Wang et al., 2018).

The use of turbidity (Turb) as single surrogate to predict other water quality parameters, mainly suspended solids (SS) and total P (TP), through data mining or machine learning methods has been addressed in recent years by numerous research works (Jones et al., 2011; Steffy and Shank, 2018; etc.). However, these works are mainly focused on linear relationships. In general, TP in rural environments can be related to the concentration of SS, which at the same time is highly correlated with Turb. Urban environments tend to be more complex which makes the occurrence of linear relationships infrequent (Viviano et al., 2014). Studies including multiple potential surrogates are scarcer (Schilling et al. (2017); Valkama and Ruth (2017)).

In spite of the interest in measuring TP, its determination is not enough to perform a proper analysis of P loads, especially when it has to do with the causes of eutrophication. TP is composed of different fractions of P, mainly the dissolved plus the particulate fraction, including both organic and inorganic components and non-bioavailable fractions. In addition, the percentage of bioavailability is highly variable in time and space due to the influence of dynamic physical-chemical processes (Renjith et al., 2011, Lannergård et al. (2019)).

Non-linear machine learning models present some advantages in comparison to linear regression. There is no need to find obvious relationships between the target variable and the surrogates, and they allow capturing phenomena that have a non-linear behaviour, as usually occurs with environmental processes. Due to the complexity of these relationships, the selection of the variables to take part in the model is not a trivial task. In the field of water sciences, non-linear data-driven models have been mainly applied for dissolved oxygen (DO) prediction; for example Olyaie et al. (2017) compared the use of two types of Artificial Neural Networks, Linear Genetic Programming and Support Vector Machine for DO prediction with on-line measurable parameters as inputs. The use of tree based models to estimate suspended sediment concentrations was reported by Francke et al. (2008), who found as main advantages the possibility to account for multiple interactions of predictors without prior knowledge and the possibility to interpret the results in case of simple interactions. According to Corominas et al. (2018), decision trees and fuzzy logic are examples of successful models in practice, from the point of view of their transferability to the end-user.

The main objective of this work was to assess the feasibility of non-linear machine learning models and their advantages in relation to linear models, for developing soft-sensors for nutrients concentrations. Since the investment and maintenance cost of soft-sensors will depend on the number and type of surrogate sensors as well as on the complexity of the models, the procedure to select the best set of variables was investigated. The output of the models, the forecasted target variables datasets, is valid for direct use or to feed process-based water quality models, like eutrophication models, in order to contribute to a holistic knowledge of river basins, improving the cost and time effectiveness of traditional sampling methods.

**2. Materials and methods**

In this work, the implementation of non-linear machine learning models to predict nutrient concentrations, using easy to measure variables as predictors, is compared to the use of linear models. A previous process of variable selection to determine the smallest set of sensors that would be needed without compromising the accuracy of the models is applied. The data (further described in Section 2.1) from the monitoring of two different catchments (River Enborne and The Cut, England) are used to train, test and validate the models. This process was carried out using seven physico-chemical predictors in each catchment (electrical conductivity (EC), Turb, temperature (Temp), DO saturation concentration (DOsat), pH, chlorophyll concentration (Chl) and the flow rate (Flow)), to predict Total Reactive Phosphorus (TRP) and nitrogen as nitrate ($NO_3$-N) concentrations in River Enborne and TRP, TP and nitrogen as ammonia ($NH_4$-N) in The Cut. A scheme of the general outline of the process followed from the acquisition of the dataset to the validation of the models is given in Figure 1.

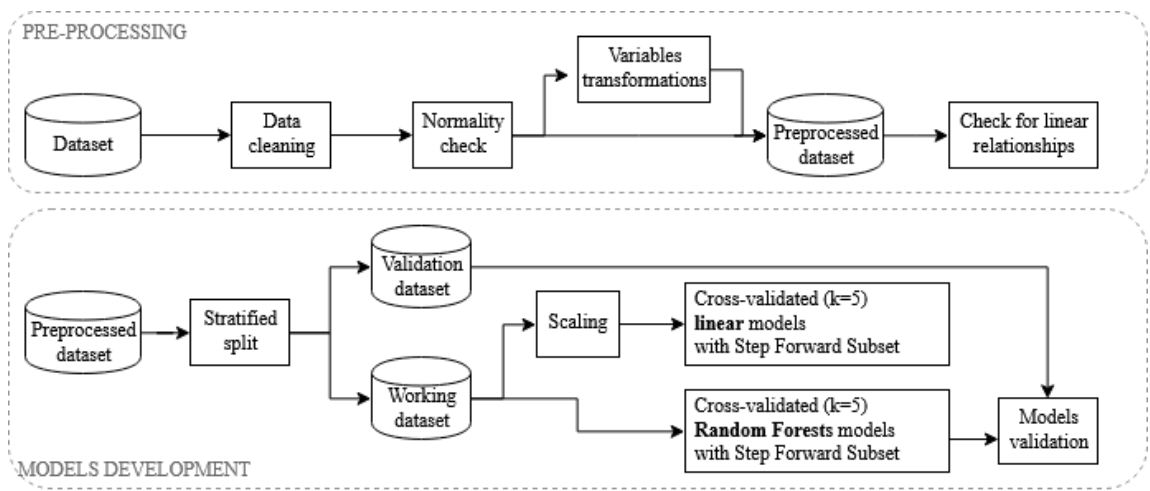

Figure 1. General overview of the process followed, from the acquisition of the dataset to the validation of the models.

**2.1. Description of the study-site and the data**

This work employs an open dataset that was obtained from the Environmental Information Data Centre platform (http://eidc.ceh.ac.uk/) of the Centre for Ecology and Hydrology (CEH) belonging to the United Kingdom's Natural Environment Research Council. For the purpose of this study, the datasets of two tributaries of the River Thames (England) with contrasting land uses were selected. The time-series data for The Cut (Bowes et al., 2015a) and River Enborne (Bowes et al., 2015b) provide hourly physical and chemical monitoring data taken between 2009 and 2012 using automated in-situ instrumentation. Detailed documentation including maps, photos and relevant information about the monitoring methodology and the instrumentation characteristics is provided with the datasets as supporting information. Additional details about the catchments and the data validation procedure with weekly grab sampling and laboratory analysis is given in Wade et al. (2012). The River Enborne is relatively rural, receiving sewage effluent from small sewage treatment works, whereas The Cut is a highly urbanised catchment receiving large quantities of sewage effluent from major towns.

The parameters measured in the River Enborne are TRP, $NO_3$-N, EC, Turb, Temp, DOsat, pH and Chl. The parameters measured in The Cut are the same as those from the River Enborne, except for dissolved N, which was measured as ammonia. In addition, TP was measured in The Cut. The accompanying hourly averaged Flow data from the closest UK's Environmental Agency gauging station are also supplied in both cases as part of the datasets.

### 2.2. Data analysis framework

The datasets are available in Comma-Separated Values format (CSV). Statistical operations and data analysis were performed using open source libraries and frameworks for the Python 2.7, which was chosen due to its growing usage trend for data analytics, being also suitable as a general purpose programming language (Nguyen et al., 2019). The main Python libraries that were used are Numpy (1.15.4), SciPy (1.1.0), Pandas (0.23.4), Scikit-Learn (0.20.1) (Pedregosa et al., 2011) and MLxtend (0.13.0) (Raschka, 2018). The specific functions used are indicated in the text where relevant.

### 2.3. Data pre-processing and exploratory analysis

The available data are already curated by the CEH and the University of Reading (Wade et al., 2012). The Enborne dataset had 20,412 records and The Cut dataset had 15,636. After removing missing values, represented as Not a Number (NaN) records, they resulted in 12,723 and 8934 records respectively. NaN records are attributed to maintenance and calibration procedures, that were carried out with different periodicity and duration depending on the probe or sensor, and due to the removal of anomalous points during the data curation process (Wade et al., 2012). A summary of water quality parameters statistics for both datasets is shown in Table 1.

|  |  | Enborne (n = 12,723) | | | The Cut (n = 8934) | | |
|---|---|---|---|---|---|---|---|
| **Parameter** | **Unit** | **Mean ± sd** | **Min** | **Max** | **Mean ± sd** | **Min** | **Max** |
| **Flow** | m$^3$ s$^{-1}$ | 0.97 ± 1.32 | 0.11 | 12.60 | 0.24 ± 0.30 | 0.06 | 4.64 |
| **Temp** | ºC | 10.85 ± 4.41 | 0.30 | 19.80 | 13.48 ± 4.00 | 2.20 | 22.9 |
| **pH** | - | 8.00 ± 0.22 | 7.24 | 8.94 | 7.72 ± 0.13 | 7.23 | 8.39 |
| **DOsat** | % | 90.28 ± 15.58 | 53.00 | 170 | 76.04 ± | 27.00 | 176.00 |
| **Turb** | NTU | 8.05 ± 10.28 | 2.00 | 215.8 | 15.47 ± | 0.70 | 1255.40 |
| **EC** | µS cm$^{-1}$ | 486.53 ± 79.64 | 215 | 769 | 958.88 ± | 367.00 | 1115.00 |
| **Chl** | µg L$^{-1}$ | 3.30 ± 1.33 | 0.10 | 17.4 | 4.51 ± 4.76 | 0.80 | 92.9 |
| **TRP** | mg P L$^{-1}$ | 0.17 ± 0.10 | 0.00 | 0.58 | 0.59 ± 0.22 | 0.15 | 1.45 |
| **TP** | mg P L$^{-1}$ | - | - | - | 0.64 ± 0.21 | 0.22 | 1.49 |
| **NO$_3$-N** | mg N L$^-$ | 4.06 ± 0.75 | 1.70 | 6.25 | - | - | - |
| **NH$_4$-N** | mg N L$^-$ | - | - | - | 0.60 ± 0.24 | 0.03 | 1.43 |

Table 1. Mean, standard deviation (sd), minimum value (Min) and maximum value (Max) of the variables contained in each dataset.

The predictors presented different scales and distributions. Normality was checked by means of Q-Q plots (represented in Annex, section A.1). Those attributes which were not normally distributed were transformed to fulfil this condition: those that presented a positive skew (Turb, Chl, DOsat and Flow in River Enborne and Turb, Chl and Flow in The Cut) were natural log transformed while EC in The Cut, which presented negative skew, was transformed by cubing.

The presence of a linear relationship between pairs of variables was checked computing the standard correlation coefficient or Pearson's *r*, as a descriptor of the degree of linear association between two variables, by means of the `pandas` method `corr()` which implements Equation 1:

$$r = \frac{\Sigma(x_i - \bar{X})(y_i - \bar{Y})}{\sqrt{\Sigma(x_i - \bar{X})^2 \Sigma(y_i - \bar{Y})^2}} \qquad \text{Equation 1}$$

where *i* represents each observation, *x* is the first variable and $\bar{X}$ is its mean value, and *y* is the second variable with $\bar{Y}$ being its mean value.

Finally, in order to train, test and validate the models, the datasets needed to be partitioned into the three corresponding fractions: train, test and validation sets. The validation set was separated from the beginning and will be used to assess the performance of the final models on unseen data. It contained 10% of the data (1,273 records for the River Enborne and 894 for The Cut). In order to obtain a representative and unbiased validation dataset we applied stratified sampling using the function `StratifiedShuffleSplit` of the library Scikit-Learn. The distribution of the target variable in the validation fraction was checked by visual inspection of the histograms to assure that this fraction was representative of the overall dataset. This process resulted in a working subset and a validation subset for each target variable. The working subset was split into train and test sets in each specific model applying *k-fold cross-validation* with *k*=5. In this method the working dataset is split into *k* fractions or folds, each one being used iteratively as a test set while the remaining *k-1* fractions were used as a training set. This means that in each iteration 80% of the working dataset act as training fraction (9160 records for the River Enborne and 6431 records in The Cut) while the remaining 20% act as test (2290 records for the River Enborne and 1609 records in The Cut).

### 2.4. Selection of variable predictors

In any machine learning model, it is always advisable to select a subset of relevant predictors to avoid overfitting and, therefore, enhance generalization. In addition, a less complex model, with fewer predictors, is advisable because it will consume less time and resources to train and execute and will gain in interpretability by the end-user. In a linear model, in spite of using predictors that a priori are independent, selecting variables is also necessary to avoid covariance.

In the present study we also sought the minimum best subset of variables due to practical implications. This subset determines the set of surrogate sensors that will take part in the operation of the soft-sensor in the future. The objective is to employ the minimum number of sensors to reduce the cost of purchasing, installing and maintaining the set of surrogate sensors that feeds the soft-sensor. In addition, those surrogates that take part in the best subset for several target variables can be shared by the different models. The more common surrogates, the smaller the set of surrogate sensors needed to predict several target variables at the same time. Diverse variable selection methods have been previously applied in water sciences modelling, like Best Subset Selection, Hierarchical Partitioning and Stepwise Selection with bootstraping (Francke et al., 2008). In this work, Stepwise Forward Subset was performed, which assesses the performance of the models with each individual variable and iteratively adds the variable that produces the highest improvement according to a certain evaluation criterion or scoring: the adjusted $R^2$ in the case of the linear models and RMSE in the case of non-linear models. In contrast to traditional $R^2$, the adjusted $R^2$ accounts for a penalization when non meaningful variables are added. It was calculated as in Equation 2.

$$\text{Adjusted } R^2 = 1 - \frac{N-1}{N-p-1}(1 - R^2) \qquad \text{Equation 2}$$

where N is the number of observations and p is the number of variables.

The RMSE was calculated as in Equation 3.

$$\text{RMSE} = \sqrt{\frac{\sum_{i=1}^{N}(y_i - \hat{y}_i)^2}{N}} \qquad \text{Equation 3}$$

where $y$ is the observed value, $\hat{y}$ is the predicted value and N is the number of observations.

### 2.5. Linear models

If a linear relationship is expected to exist between the target variable and the predictors, a Multiple Linear Regression (MLR) model can be used, expressed as in Equation 4.

$$y_i(w, x_i) = w_0 + w_1 x_{i,1} + \ldots + w_n x_{i,n} \qquad \text{Equation 4}$$

$y_i$ is the predicted value, $w$ is the vector that contains the coefficients with $w_0$ being the intercept and $x_i$ is the i$^{th}$ observation of the dataset with $n$ attributes.

The Sequential Feature Selector of the library `mlxtend` was implemented to select the subsets of one to seven predictors and fit the corresponding linear models, using the adjusted $R^2$ as a reference scoring criterion and after scaling the predictors to zero mean and unit variance. The models were cross-validated with $k=5$. Finally, the RMSE of the models with one to seven predictors was calculated to be able to compare their performance with that of non-linear models.

### 2.6. Non-Linear models

The criteria taken into account to select the type of non-linear models were mainly the suitability of the model to the available data and the usability of the model by the end-user. Decision trees were chosen because they are easy to interpret in comparison with other machine learning techniques, as the way in which they split the data space is similar to the way humans would do. Data do not need to follow any special distribution, and these models are not very sensitive to outliers and noise. Since trees only consider one predictor at each partition, the comparative scales of the predictors do not influence the model. These characteristics make decision trees very suitable to deal with the varied typology and temporal variability of water quality data.

Trees tend to overfit the model to the training sample (Breiman et al., 1984), but by aggregating many decision trees the predictive performance can be substantially improved. The well-known machine learning algorithm called Random Forests (RF) (Breiman, 2001) averages the result of many independent individual trees with a random subset of data and predictors in order to mitigate the poor performance of any single tree. A comprehensive review of its implementation in the water science field has been recently published (Tyralis et al., 2019). Details about the implementation of decision trees in this work are given in the Annex (section A.2), in case the reader is not familiar with RF.

In this work, the `RandomForestRegressor` function of the library Scikit-Learn was used, which fits a number of regression trees on various sub-samples of the dataset and uses averaging to improve the predictive accuracy and control over-fitting. To find out the ideal value of the hyperparameters, the models were run on a predefined grid of values. This process was performed applying the function `GridSearchCV` of Scikit-Learn, using cross-validation with $k = 5$ and 50 trees (a low number of trees to speed up the searching process). The hyperparameters and their values contained in the grid are shown in Table 2. These gave place to 162 combinations, that multiplied by the 5 folds resulted in 810 fits. Once the hyperparameters were fitted, the number of trees was adjusted by analysing the performance of the model when using 1, 10, 50, 100 and 200 trees. The highest number of trees that provided an improvement of more than 5% on the test error with respect to the previous number of trees was chosen.

In order to perform the variables subset selection, a procedure similar to that employed in the linear method was implemented here, but in this case the reference scoring was the cross-validated ($k = 5$) RMSE. In this case, the pre-processing was simpler because no variable transformation or scaling is needed in RF models. In order to avoid the influence of some variables over the rest, the feature selection was performed recursively, that is to say, in the first round the variable that was included in the last place was removed. Then, in a second round, a new model was fitted on the new predictors subset, and again the last variable was removed. This was repeated until just one variable remained in the subset.

| Hyperparameter | Values |
| --- | --- |
| Bootstrap | True, False |
| Size of the random subsets | auto[1], sqrt[2], |
| Depth of the trees | 10, 20, 30 |
| Minimum number of samples to split a node | 6, 12, 20 |
| Minimum number of samples to be at a leaf node | 6, 12, 20 |

Table 2. Values of the hyperparameters contained in the grid search.

[1] The size of the random subsets of predictors to consider when splitting a node is the total number of predictors in the model or [2] the square root of the total number of predictors in the model.

Finally, the selected models were validated using the previously split validation dataset. Besides the RMSE, a relative error metric was calculated: the normalised RMSE (nRMSE) which is the result from dividing the RMSE by the mean of the observed data. While the RMSE has the units of the measured variable, the nRMSE is expressed in percentage, which allows comparing variables with different units.

In order to check whether the final models tended to underestimate or overestimate the forecasts, the percentage bias (PBIAS) was calculated as in Equation 5.

$$PBIAS\ (\%) = \frac{\Sigma(\hat{y}_i - y_i)}{\Sigma(\hat{y}_i)} 100 \qquad \text{Equation 5}$$

If the value of PBIAS is 0 the model is unbiased, while positive and negative values represent overestimation and underestimation respectively.

## 3. Results and discussion

### 3.1. Exploratory analysis

In general, the predictors did not present strong correlations with the target variables (r < 0.76), as shown in Table 3. Correlations turned out to be stronger in the rural catchment than in the urban one for nearly all the predictors in relation with the targets. In all the relationships reported here, the significance level is under 0.01, which is attributed to the relatively large size of the datasets. Therefore, the relationships are considered highly significant.

| River Enborne | | | | The Cut | | | | | |
|---|---|---|---|---|---|---|---|---|---|
| **TRP** | | **NO$_3$-N** | | **TRP** | | **TP** | | **NH$_4$-N** | |
| EC | 0.72 | EC | 0.76 | EC | 0.44 | EC | 0.44 | Temp | 0.29 |
| Flow | -0.67 | Temp | 0.55 | Flow | -0.30 | Flow | -0.29 | Chl | -0.17 |
| DOsat | -0.60 | Flow | -0.54 | DOsat | -0.22 | DOsat | -0.22 | Turb | 0.16 |
| Temp | 0.52 | Chl | -0.33 | Turb | -0.15 | Turb | -0.12 | Flow | -0.16 |
| Chl | -0.37 | DOsat | -0.31 | Chl | -0.12 | Temp | 0.10 | EC | 0.12 |
| Turb | 0.11 | pH | 0.16 | Temp | 0.08 | Chl | -0.09 | DOsat | 0.02 |
| pH | -0.11 | Turb | -0.06 | pH | 0.02 | pH | 0.01 | pH | 0.00 |

Table 3. Pearson's *r* for each of the target variables with respect to the seven predictors, ordered by the absolute value of *r*.

Considering TRP as the target variable, in the River Enborne the Pearson's coefficient reveals a strong linear relationship with EC (r = 0.72), Flow (r = -0.67) and DOsat (r = -0.60). Regarding NO$_3$-N, it also shows a strong linear relationship with EC (r = 0.76) and some weaker with Temp (r = 0.55) and Flow (r = -0.54). With the rest of the variables the relationship is weak. In spite of being a relatively rural catchment, 67% of TP load is as soluble form (Bowes et al., 2015c), thus Turb is not useful as main surrogate to predict TRP because it seems that P is nor associated to particulate material. The relationship between nutrient concentrations and Flow is inverse, which is in agreement with the seasonal pattern of nutrients dynamics in River Enborne. High nutrients concentrations coincide with low flow periods in summer and autumn, while low nutrients concentrations take place during high flow winter and spring periods (Bowes et al., 2015c). This seasonal behaviour also explains the strong positive relationship between nutrients concentrations and Temp. Nutrients concentrations are higher in summer and autumn coinciding with high temperature periods, while low nutrients concentrations coincide with low temperature periods.

NH$_4$-N in the Cut does not present a strong linear relationship with any variables. The highest *r* is 0.29, and it occurs with Temp. In the case of TRP and TP, the strongest correlation occurs with EC (r = 0.44). Even the relationship between TP and Turb is weak (r = -0.12). Besides, it is negative, which suggests that the majority of the phosphorus is found in dissolved form, or at least not associated with particles that could be detected by means of the measurement of Turb. This is in agreement with that reported by Halliday et al., (2015), which indicates that soluble reactive P (SRP) is the dominant P fraction, accounting for on average 78% of the TP in The Cut catchment, where 50.3% of the annual P input loads come from human sewage effluent discharges. According to Halliday et al., (2015), the effluent of the sewage treatment facility has a great influence on the water quality dynamics. This could explain the absence of strong relationships with other environmental parameters, like temperature. In addition, important phytoplankton blooms have not been detected in The Cut (Halliday et al., 2015), attributed to the reduced water residence times. This reduced the influence of biological processes on nutrient dynamics.

From a practical point of view, EC presents the strongest correlation with four of the five target variables analysed, while pH presents the weakest one also in four of five cases. Turb, which is considered a promising surrogate for TP prediction (Schilling et al., 2017), presents a low correlation coefficient with the five target variables (r ≤ 0.16), which means that it would not be a good predictor by itself but could bring information along with other variables. Flow is an important variable because it is in the top-three for four of five target variables. However, its inclusion in the models will depend on the degree of covariance with EC.

### 3.2. Linear models

The first step to obtain the linear models was the selection of the predictors, according to the procedure described in Section 2.4. The performance of the models with one to seven predictors, in terms of the adjusted $R^2$ for both training and testing fractions is shown in Figure 2A and 2B. The variables used in each case are shown in Table 4. The models fitted on training data had a very similar performance with test data; therefore, no further regularization was applied. This is attributed to the high ratio between number of observations and number of predictor variables.

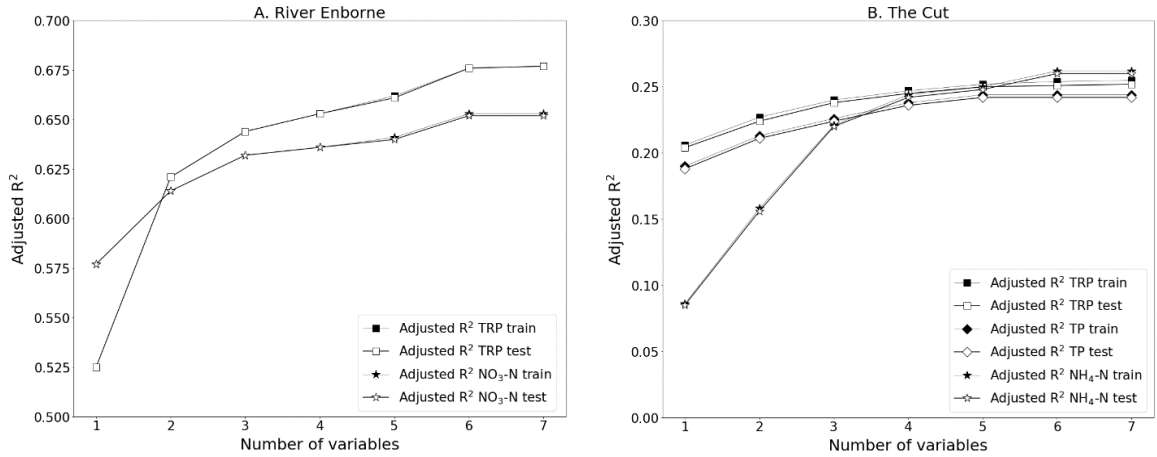

Figure 2. Adjusted $R^2$ of the MLR models as a function of the number of variables in River Enborne (A) and The Cut (B). In order to make the visualization of the results easier, the scale of the adjusted $R^2$ axis is different for the graphs A and B.

The performance of the linear models, based on the value of the adjusted $R^2$, is notably better in the rural catchment than in the urban one. In River Enborne, a $R^2$ higher than 0.50 was achieved just using one predictor (CE) and continues improving until six predictors are used, achieving an $R^2$ of 0.68 and 0.65 to predict TRP and $NO_3$-N respectively. On the contrary, in The Cut the highest $R^2$ was 0.26 with six predictors to predict $NH_4$-N. The predictions of TRP and TP exhibit very low variations among the models with one to seven predictors. The highest observed $R^2$ values are 0.25 and 0.24 to predict TRP and TP with four and five predictors respectively. These results reveal that the linear models are more useful in the rural catchment than in the urban one, in agreement with previous findings where the complexity of urban watersheds is highlighted. (Viviano et al., 2014).

The order of inclusion of the variables in the models along with the test RMSE are shown in Table 4. As expected, the order of inclusion of the variables in the linear models does not coincide with the ranking according to the absolute value of Pearson's *r* (Table 1), except for the first variable: Temp in the case of $NH_4$-N in The Cut and EC for the rest of the cases. This difference in the order of the variables is due to the occurrence of some degree of covariance: those variables that provide a high amount of redundant information to the model are left for the end of the selection. In contrast, other variables being less correlated with the target variable may be included in the subset because they explain a part of the information that is not explained by the most correlated variables. This is the case with pH, that in spite of presenting very low correlation with the five targets ($0.16 \geq r \geq -0.11$), appears in the top-three in three of five targets according to its contribution to a MLR.

The occurrence of covariance between the different variables has relevant practical implications that must be taken into account in the soft-sensor design process. If the set of surrogate sensors contains a main sensor that provides the highest part of information and several non-covariant sensors, one can obtain much information by making use of just a few sensors. However, in the case of absence of data from the main sensor (i.e., out of use due to calibration, maintenance, drifting, failure, etc.) the performance of the soft-sensor would be seriously affected because the rest of the sensors provide complementary, but not redundant, information.

| Order of | 1st | 2nd | 3rd | 4th | 5th | 6th | 7th |
|---|---|---|---|---|---|---|---|
| **TRP in River Enborne** | | | | | | | |
| **Linear model** | EC | + pH | + Turb | + Chl | + Flow | + DOsat | + Temp |
| **Test RMSE** | 0.067 | 0.059 | 0.058 | 0.057 | 0.056 | 0.056 | 0.055 |
| **RF model** | EC | + Flow | + Temp | + Turb | + pH | + DOsat | + Chl |
| **Test RMSE** | 0.063 | 0.045 | 0.035 | 0.027 | 0.024 | 0.023 | 0.025 |
| **NO$_3$-N in River Enborne** | | | | | | | |
| **Linear model** | EC | + Flow | + Chl | + Temp | + pH | + DOsat | + Turb |
| **Test RMSE** | 0.487 | 0.464 | 0.454 | 0.451 | 0.448 | 0.441 | 0.441 |
| **RF model** | EC | + pH | + Flow | + Temp | + DOsat | + Turb | + Chl |
| **Test RMSE** | 0.459 | 0.348 | 0.253 | 0.174 | 0.165 | 0.161 | 0.168 |
| **TRP in The Cut** | | | | | | | |
| **Linear model** | EC | + DOsat | + pH | + Turb | + Flow | + Chl | + Temp |
| **Test RMSE** | 0.194 | 0.191 | 0.189 | 0.188 | 0.188 | 0.188 | 0.188 |
| **RF model** | EC | + Turb | + Temp | + pH | + Flow | + DOsat | + Chl |
| **Test RMSE** | 0.191 | 0.183 | 0.141 | 0.122 | 0.113 | 0.112 | 0.106 |
| **TP in The Cut** | | | | | | | |
| **Linear model** | EC | + DOsat | +pH | + Turb | + Flow | + Chl | + Temp |
| **Test RMSE** | 0.190 | 0.187 | 0.186 | 0.184 | 0.184 | 0.184 | 0.184 |
| **RF model** | EC | + Turb | + Temp | + pH | + Flow | + DOsat | + Chl |
| **Test RMSE** | 0.194 | 0.169 | 0.143 | 0.118 | 0.106 | 0.105 | 0.104 |
| **NH4-N in The Cut** | | | | | | | |
| **Linear model** | Temp | + Turb | + Chl | + EC | + pH | + DOsat | + Flow |
| **Test RMSE** | 0.228 | 0.219 | 0.211 | 0.208 | 0.207 | 0.205 | 0.205 |
| **RF model** | Chl | + Temp | + Turb | + pH | + EC | + Flow | + DOsat |
| **Test RMSE** | 0.214 | 0.189 | 0.149 | 0.121 | 0.102 | 0.116 | 0.098 |

Table 4. Test RMSE for each target variable and each subset of variables of size one to seven.

### 3.3. Non-linear models

#### 3.3.1. Performance of the models

As a general rule, the RF models outperformed the MLR models except when just one predictor variable was used, where the results were practically the same with both types of models. As predictors were added, the RF models resulted in considerably lower RMSE than the MLR models. This means that, with the data of this study, RF obtain much more benefit from introducing new predictors than MLR. The values of percentage of reduction of test RMSE as predictors are added in both linear and non-linear methods are presented in Figure 3. When the seven variables were used, the reduction of the RMSE in relation to using just one predictor in RF ranged from 44.5% to 63.4%, while in MLR the highest improvement was 17.9%.

According to this work, if just one surrogate is available, implementing RF is not worth the effort in comparison with MLR. Adding another surrogate generates some improvement, but when RF really outperforms MLR is when at least three predictors are used. Then there exists an optimum from which adding more predictors does not improve the performance of the models. For the River Enborne, the RF models with more than four variables did not improve their test RMSE in more than 5%. The same occurred with the RF models of The Cut with more than five variables. With four surrogates in River Enborne, the RMSE of the predictions of TRP and $NO_3$-N are reduced to 47.4% and 38.6% respectively, in comparison with linear models. With sets of five sensors in The Cut, the RMSE of the predictions of TRP, TP and NH4-N were reduced to 60.1%, 57.6% and 49.3% respectively in comparison with linear models.

RF demonstrates to be very efficient exploiting the available information which highlight their suitability to represent environmental phenomena in contrast to linear models, which are not able to exploit much of the information that can be collected by making use of simple and highly available sensors.

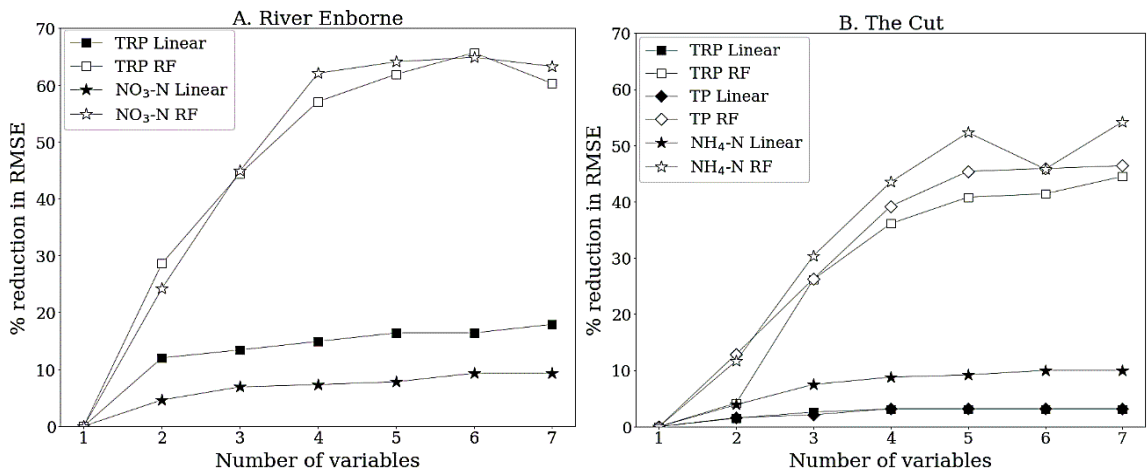

Figure 3. Percentage of reduction of test RMSE as a function of the number of predictors in the model for the linear and the RF models in River Enborne (A) and The Cut (B).

### 3.3.2. Order of inclusion of the variables

As observed in the linear models, the first predictor is, with the exception of the model for $NH_4$-N, the one with the highest Pearson's *r* with respect to the target variable. According to Tables 3 and 4, the order of inclusion of additional predictors does not follow a specific pattern. This is an important fact to take into account when building machine learning models. The complex relationships that lie behind these models result in the utilization of predictor variables that a priori are not relevant, neither intuitively nor upon analysis of correlation coefficients or stepwise subsets performed with linear models. According to the results of the present work, using a procedure in accordance to the model that is going to be implemented is crucial to perform the variable subsetting process. Methods like those based on feature weight coefficients or feature importance may be less complex and less resources consuming; however, selecting variables based on user-defined performance metrics performs better when the objective is to select the minimum number of predictors without compromising the error of the model.

In River Enborne the best subsets of four surrogates for the two predicted substances only differ in one surrogate. Both models use EC, Flow and Temp, but Turb is used to predict TRP and pH is used to predict $NO_3$-N. This means that a set of five surrogate sensors would have the ability to predict TRP and $NO_3$-N at the same time with the calculated accuracy. In the case of The Cut, the models to predict TRP and TP incorporate identical predictors: EC, Turb, Temp, pH and Flow. This subset only differs in one surrogate in relation to the model to predict $NH_4$-N, which includes Chl instead of Flow. In practice, a subset of six sensors could be used to predict the three target variables at the same time.

### 3.3.3. Performance on unseen data

The RF models with up to four predictors in River Enborne and with up to five variables in The Cut were run on the validation dataset to assess the RMSE, the nRMSE and the PBIAS on unseen data. RMSE and nRMSE values are shown in Table 5. This kind of table would allow the water managers, or other stakeholders, to make a cost-benefit analysis to choose the number of sensors to install as a function of the achievable RMSE.

| Predictors in the RF models | RMSE ± sd | nRMSE (%) |
|---|---|---|
| **TRP in River Enborne** | | |
| EC | 0.061 ± 0.043 | 35.69 |
| EC, Flow | 0.043 ± 0.035 | 24.92 |
| EC, Flow, Temp | 0.030 ± 0.025 | 19.22 |
| EC, Flow, Temp, Turb | 0.025 ± 0.021 | 15.47 |
| **$NO_3$-N in River Enborne** | | |
| EC | 0.458 ± 0.286 | 11.40 |
| EC, pH | 0.343 ± 0.229 | 7.81 |
| EC, pH, Flow | 0.254 ± 0.186 | 6.22 |
| EC, pH, Flow, Temp | 0.194 ± 0.138 | 4.78 |
| **TRP in The Cut** | | |
| EC | 0.199 ± 0.115 | 32.71 |
| EC, Turb | 0.180 ± 0.112 | 28.54 |
| EC, Turb, Temp | 0.141 ± 0.099 | 23.78 |
| EC, Turb, Temp, pH | 0.117 ± 0.088 | 20.46 |
| EC, Turb, Temp, pH, Flow | 0.107 ± 0.078 | 18.34 |
| **TP in The Cut** | | |
| EC | 0.192 ± 0.114 | 29.95 |
| EC, Turb | 0.173 ± 0.116 | 26.14 |
| EC, Turb, Temp | 0.153 ± 0.104 | 21.74 |
| EC, Turb, Temp, pH | 0.119 ± 0.087 | 18.55 |
| EC, Turb, Temp, pH, Flow | 0.110 ± 0.081 | 16.93 |
| **$NH_4$-N in The Cut** | | |
| Chl | 0.210 ± 0.144 | 34.51 |
| Chl, Temp | 0.190 ± 0.123 | 26.17 |
| Chl, Temp, Turb | 0.150 ± 0.104 | 23.77 |
| Chl, Temp, Turb, pH | 0.128 ± 0.095 | 20.88 |
| Chl, Temp, Turb, pH, EC | 0.107 ± 0.077 | 19.22 |

Table 5. Validation RMSE ± standard deviation (sd) (units of the measured variable) and nRMSE (%) of the selected RF models for each of the five target variables.

The values of PBIAS were between -0.90% and 0.95%, which means that the models are unbiased (results shown in Annex, section A.3).

In Figure 4 the observed and predicted values for the first 100 observations of the validation dataset are represented for each of the five target variables and for both types of models. The models that have been used to generate these plots are the models with four variables in the case of River Enborne and with five variables in the case of The Cut.

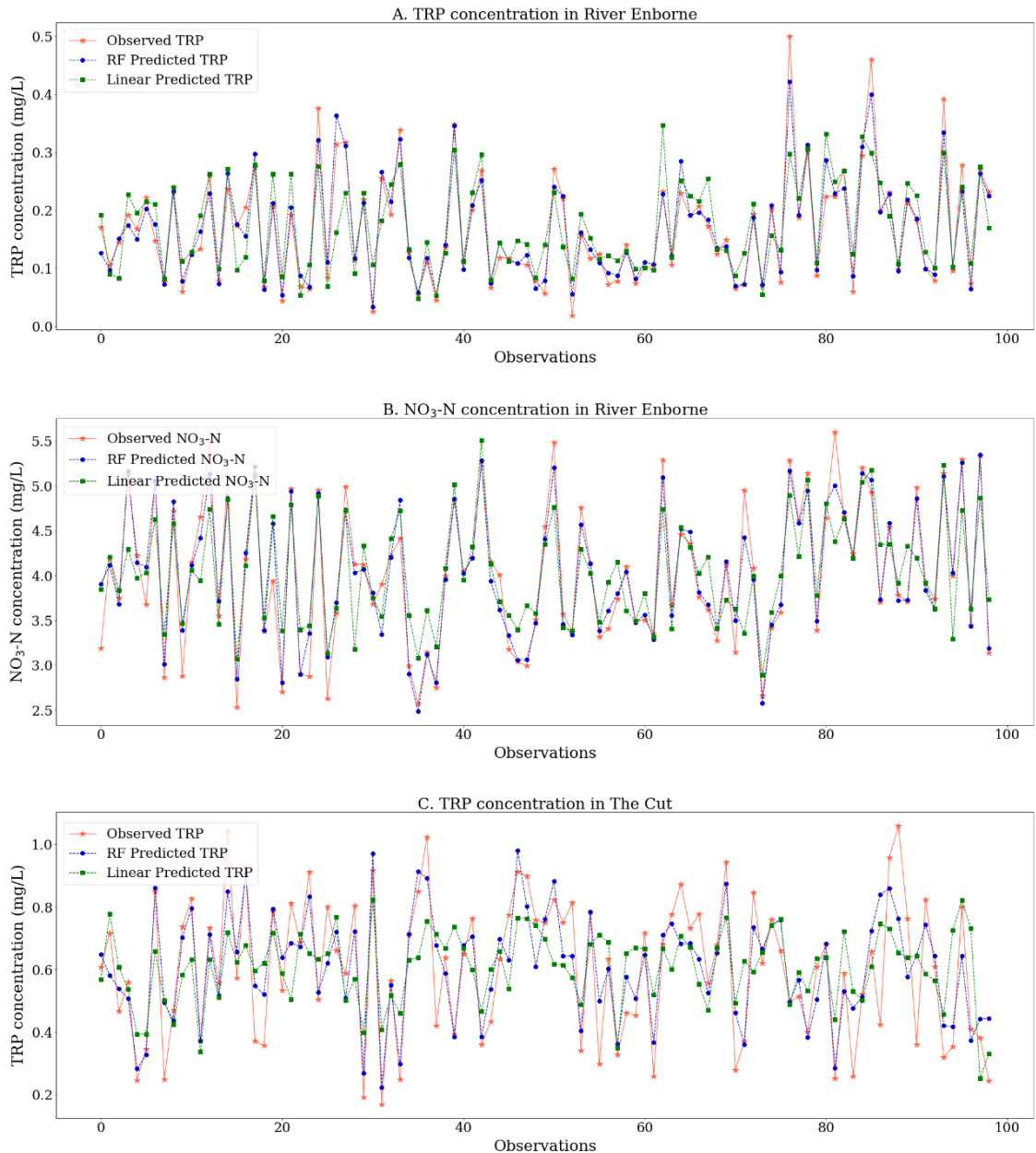

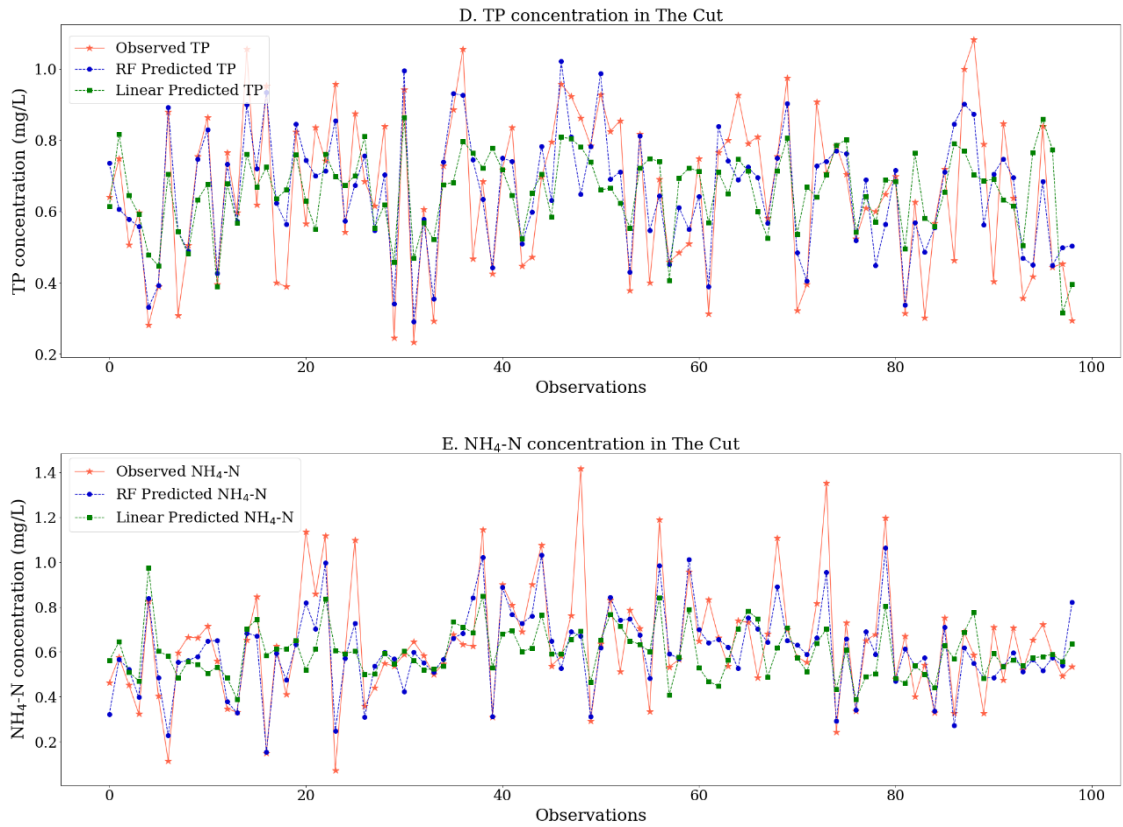

Figure 4. Observed and predicted values for the first 100 observations of the validation dataset for **A**. TRP in River Enborne, **B**. $NO_3$-N in River enborne, **C**. TRP in The Cut, **D**. TP in the Cut and **E**. $NH_4$-N in The Cut.

### 3.3.4. Hyperparameters and architecture of the models

The selected hyperparameters during the grid search process determined the architecture of the optimum models for each target variable and each number of predictors. The complete list of the selected hyperparameters is shown in the Annex (section A.2). Models containing four predictors or more performed better without bootstrapping observations but selecting just a fraction (the squared root) of the predictors when splitting a node. On the contrary, when three predictors or less were used, the trees were built performing bootstrapping on the observations but the predictors were not sampled. This means that in the case of a low number of predictors (three or less in this case), RF introduces the randomness through the sampling of the observations. On the contrary, when the number of predictors is higher the randomness is introduced through the sampling of the predictors when splitting each node.

In addition, forests with three predictors or less are shallower, and terminal nodes or leaves tend to be bigger than those of trees with four predictors or more. These results are consistent with the idea of avoiding over-fitting. The fewer the predictors, the less splits need to be done.

To illustrate the appearance of the trees, an extract of one of the 100 trees of the RF model to predict TRP in River Enborne is depicted in Figure 5. As described before, this model has four predictors: EC, Flow, Temp and Turb. In this tree, the input space is split according to the value of Flow in the root node. In a second level, the observations that satisfies the condition of the root node are split according to the value of Turb while the rest are split according to the EC. In this model, in each split the squared root of the predictors is considered, which means that the model chooses between two predictors when looking for the best split. According to Figure 5, in the fifth level there are three leafs or terminal nodes that contain 6, 8 and 10 observations. Splitting these nodes would have given place to terminal nodes with less than 6 observations, but according to the hyperparameter that defines the minimum number of observations to take part of a leaf node, whose value is 6, this is not possible. The other internal nodes in the fifth level, continue splitting until they meet a stop criterion.

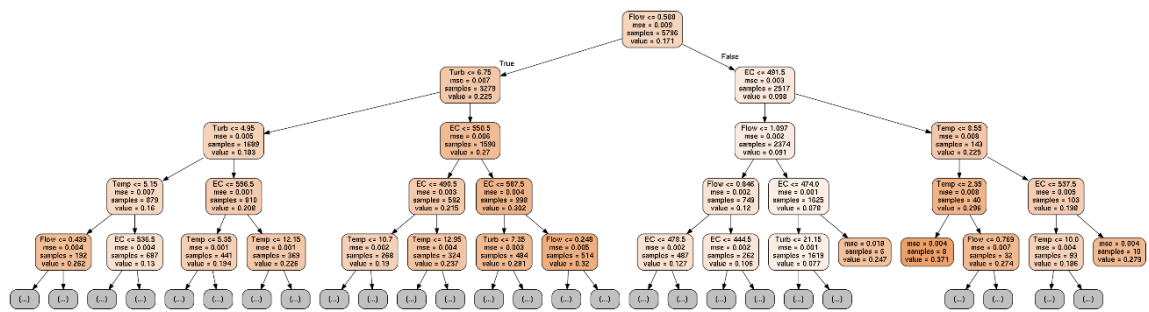

Figure 5. Illustrative example of an extract of one of the 100 trees that takes part of the RF model to predict TRP in River Enborne. *MSE* stands for Mean Squared Error, *samples* indicate the number of observations included in each node and *value* is the mean value of the target variable for the training observations that belong to each node.

### 3.3.5. Use of shared open data

The fitted models can be used for the development of soft-sensors for nutrient concentrations. If the soft-sensors are provided with a network connection, they can benefit from using information that is already measured by other institutions or organizations and is freely available on the Internet. This allows including additional information to the models without bearing the cost of the sensors and their maintenance. Nowadays a large number of water agencies provide real or quasi real time information in their Automatic Hydrological Information Systems (AHIS). In fact, in the datasets used in this work, the variable Flow comes from the closest UK's Environmental Agency flow gauging station.

Considering that the flow rate is a variable that is usually available and provided by AHIS, it is worth assessing its potential as a predictor. In fact, it appears on the second place in the subset to predict TRP in River Enborne and in the third place to predict $NH_4$-N. Its usefulness in The Cut is expected to be much lower, because it appears in the fifth place to predict TRP and TP and does not appear in the subset selection to predict $NH_4$-N. In these cases, however, Flow could be used as a backup in case of failure of the main sensor. The models were fit and run using only Flow as input and also together with the most important variable in each case. In the case of River Enborne, the most important predictor was EC. Using only Flow as predictor resulted in a validation RMSE of 0.062 and 0.602 for TRP and $NO_3$-N respectively, while adding EC resulted in test RMSE of 0.043 and 0.347 respectively, which are lower than the minimum RMSE achieved in this work with MLR and comparable to the RMSE when using the two optimal predictors for each case with RF. In the urban catchment using only Flow and another variable (EC in the case of TP and TRP and Chl in the case of $NH_4$-N) produces a slightly lower RMSE than the minimum achieved with MLR and also comparable to the RMSE when using the two optimal predictors for each case with RF.

These results highlight the existence of valuable information that can be used to obtain derived data at low cost. It is worthwhile to highlight the growing importance of taking into account existing open data services that, thanks to the development of ubiquitous and affordable network communications, can be incorporated into the models, providing additional data sources at virtually zero cost. This has a positive impact on the research community as well as on the society, because it fosters collaboration, avoids the cost of redundant experiments and increases transparency, among other advantages.

## 4. Conclusion

RF models have a great potential to reproduce environmental phenomena and can contribute to the on-line high frequency monitoring of water quality constituents.

In this work, the RF outperformed the linear models except when just one predictor was used, where the results were practically the same with both types of models.

RF showed higher ability to exploit more of the information that can be collected by making use of economical and highly available sensors, as proved by their improved performance when up to five predictors were added to the models.

The complex relationships that lie behind machine learning models resulted in the benefit of including variables that a priori were not relevant. Therefore, the variable selection process cannot be seen as an isolated process to be performed independently of the model to be applied.

The performance of both types of models, MLR and RF, was better in the rural catchment than in the urban one.

This kind of contributions demonstrates the usefulness of open data to develop research based on existing data and could incentivize the generation and promotion of new open data services and datasets to encourage the exchange of information.

The results of this work highlight the suitability of RF to represent complex environmental phenomena in contrast to linear models, which make them a promising technique for soft-sensor programming.

**Acknowledgements**

This work has been supported by the projects XDC "eXtreme-DataCloud" and DEEP-HybridDataCloud "Designing and Enabling E-infrastructures for intensive Processing in a Hybrid DataCloud" that have received funding from the European Union's Horizon 2020 Research and Innovation Programme under Grant Agreement No. 777367 and No. 777435 respectively.